\documentclass[sn-mathphys, icol]{sn-jnl}
\usepackage[utf8]{inputenc}
\usepackage{xcolor}
\usepackage{amsmath}
\usepackage{amssymb}
\usepackage{mathtools}
\usepackage{makecell}
\usepackage{amsthm}
\usepackage{float}
\usepackage{amsmath}
\usepackage{hyperref}
\usepackage{tabularx}
\usepackage{graphicx}
\usepackage{caption}
\usepackage{subcaption}
\usepackage{booktabs} 
\usepackage{algorithm,algpseudocode}
\usepackage{array}
\usepackage{enumitem}
\usepackage{lmodern}
\usepackage{cancel}
\usepackage{arydshln}
\usepackage{gensymb}

\usepackage[square,numbers,sort&compress]{natbib}
\theoremstyle{plain}

\theoremstyle{definition}

\theoremstyle{remark}

\usepackage{todonotes}
\usepackage{cleveref}

\newcolumntype{Y}{>{\centering\arraybackslash}X}

\title[]{Vision Transformers for Multi-Variable Climate Downscaling: Emulating Regional Climate Models with a Shared Encoder and Multi-Decoder Architecture}

 \author*[1,2]{\fnm{Fabio} \sur{Merizzi}}\email{fabio.merizzi@unibo.it}
 \author[2]{\fnm{Harilaos} \sur{Loukos}}

 \affil[1]{\orgdiv{Department of Informatics: Science and Engineering (DISI)}, \orgname{University of Bologna}, \orgaddress{\street{Mura Anteo Zamboni 7}, \city{Bologna}, \postcode{40126},
 \country{Italy}
 }}

 \affil[2]{\orgdiv{The Climate Data Factory (TCDF)}, \city{Paris}, \country{France}
 }

\abstract{Global Climate Models (GCMs) are critical for simulating large-scale climate dynamics, but their coarse spatial resolution limits their applicability in regional studies. Regional Climate Models (RCMs) address this limitation through dynamical downscaling, albeit at considerable computational cost and with limited flexibility.
Deep learning has emerged as an efficient data-driven alternative; however, most existing approaches focus on single-variable models that downscale one variable at a time. This paradigm can lead to redundant computation, limited contextual awareness, and weak cross-variable interactions.
To address these limitations, we propose a multi-variable Vision Transformer (ViT) architecture with a shared encoder and variable-specific decoders (1EMD). The proposed model jointly predicts six key climate variables: surface temperature, wind speed, 500 hPa geopotential height, total precipitation, surface downwelling shortwave radiation, and surface downwelling longwave radiation, directly from GCM-resolution inputs, emulating RCM-scale downscaling over Europe.
Compared to single-variable ViT models, the 1EMD architecture improves performance across all six variables, achieving an average MSE reduction of approximately 5.5\% under a fair and controlled comparison. It also consistently outperforms alternative multi-variable baselines, including a single-decoder ViT and a multi-variable U-Net. Moreover, multi-variable models substantially reduce computational cost, yielding a 29–32\% lower inference time per variable compared to single-variable approaches.
Overall, our results demonstrate that multi-variable modeling provides systematic advantages for high-resolution climate downscaling in terms of both accuracy and efficiency. Among the evaluated architectures, the proposed 1EMD ViT achieves the most favorable trade-off between predictive performance and computational cost.}

\begin{document}
\maketitle

\noindent \textbf{keywords:} Climate Downscaling, Climate Model Emulation, Vision Transformers, Multi-Task Learning, Multi-Variable Models

\markboth{}{}
\section{Introduction}\label{sec:Introduction}
Global Climate Models (GCMs) are essential tools for simulating the Earth's climate system and projecting future climate change. However, their coarse spatial resolution, typically around 1° (approximately 100 km, depending on the latitude), limits their ability to accurately capture regional-scale dynamics. To overcome this limitation, Regional Climate Models (RCMs) are employed to dynamically downscale GCM outputs, providing finer-scale climate information at resolutions typically ranging from 0.1° to 0.5°  (respectively 10 to 50 km). Although RCMs enhance spatial detail and realism, they are computationally demanding, limiting the number of possible model runs in the context of regional coordinated experiments \cite{Jacob2014}.  

An emerging alternative to traditional numerical downscaling involves the use of deep learning techniques to emulate the RCM downscaling process. Instead of performing computationally intensive simulations, deep learning models learn to reproduce the mapping from coarse-resolution GCM outputs to high-resolution RCM fields from available data. This approach presents a rapid, data-driven method for producing detailed climate projections applicable across multiple regions and scenarios, increasing the number of available simulations \cite{Rampal2024}.

\begin{figure}[h]
\centering
\includegraphics[width=0.7\linewidth]{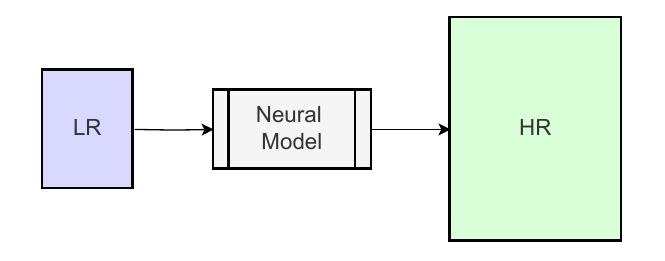}
\caption{Super-resolution problem setup. A low-resolution field (LR, left) is fed into a neural model (center), which outputs a super-resolved field (HR, right).}
\label{fig:super_resolution_setup}
\end{figure}

In this study, we frame the GCM-to-RCM emulation task as a supervised deep learning problem, as depicted in Figure \ref{fig:super_resolution_setup}. Using paired datasets from GCM and RCM outputs, we employ Vision Transformer (ViT) models to learn the relationship between coarse-resolution inputs and their high-resolution counterparts over the European domain. The primary objective of this approach is to produce spatially detailed outputs that accurately capture the fine-scale structures inherent to RCM simulations, enabling efficient and cost-effective generation of high-resolution climate data.

The main contribution of this work is the exploration of joint, multi-variable processing. Typically, downscaling meteorological variables from GCM to RCM involves handling numerous variables spanning atmospheric and land-surface categories. Most existing deep learning approaches for climate downscaling focus exclusively on single-variable modeling, training and applying uni-variate models separately on each variable \cite{bano2020configuration, vandal2017deepsd, oyama2023deep, sha2020deep, sha2020deep2}. This approach poses several issues: single-variable models cannot exploit valuable cross-variable interactions, managing and executing multiple independent models is inefficient, and similar meteorological variables result in redundant learning efforts when modeled separately. In response, this study explores multi-variable processing strategies.

\begin{figure*}[h]
\centering
\includegraphics[width=\linewidth]{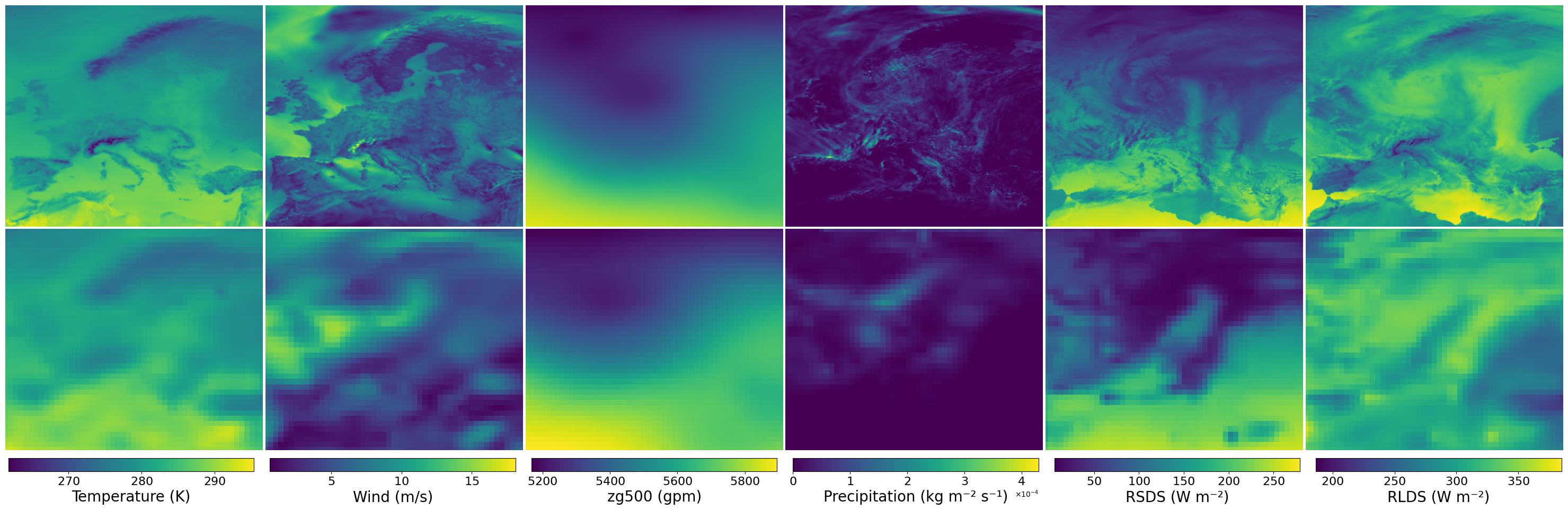}
\caption{Visualization of a single daily sample for the six studied variables: Surface Temperature, Wind Speed, Geopotential Height, Total Precipitation, Surface Downwelling Shortwave Radiation and Surface Downwelling Longwave Radiation. Each variable is shown for both the high-resolution Regional Climate Model ICTP-RegCM4-6 (top row) and the low-resolution Global Climate Model MPI-ESM-LR (bottom row) re-gridded onto the same European domain. }
\label{fig:introduction_sample}
\end{figure*}

Vision Transformers (ViTs) have emerged as powerful architectures in image-to-image translation and super-resolution tasks due to their attention-based processing and ability to capture long-range dependencies \cite{zhong2024investigating, yang2024successful, Rampal2024, liu2025spatial}. Leveraging this architecture, we propose processing several meteorological variables simultaneously, enabling the effective exploitation of inter-variable relationships and potentially enhancing overall performance and physical coherence. This approach significantly reduces the number of required models, thereby decreasing computational demands per variable and improving efficiency. However, multi-variable modeling also introduces challenges such as potential training instabilities and cross-variable interference, which may manifest as artifacts in outputs.

To evaluate multi-variable modeling strategies for GCM-to-RCM downscaling, we leverage modern ViT architectures to model this image-to-image translation problem. Our study focuses on six atmospherical variables: wind speed (sfcWind), temperature (tas), geopotential height at 500 hPa (zg500), total precipitation (tp), surface downwelling shortwave radiation (rsds) and surface downwelling longwave radiation (rlds). An image depicting a sample of the six variables for both RCM and GCM is reported in Figure \ref{fig:introduction_sample}. We first establish baseline performance using single-variable ViT models trained independently for each variable. Subsequently, we introduce two multi-variable approaches: one treating variables as separate channels (1E1D), and a second employing a single shared encoder with multiple dedicated decoders (1EMD), one per variable. In addition, we incorporate a multi-variable U-Net model to provide an external and widely used baseline for comparison. We ensure a robust and reproducible comparison by standardizing model structure, parameter count, training epochs, learning rates, and optimizers across all experiments.

Our findings reveal that multi-variable modeling offers clear advantages in terms of computational efficiency. However, the simpler multi-channel approach (1E1D) suffers from cross-variable interference, failing to match the performance of single-variable baselines. In contrast, our structured multi-decoder design (1EMD) avoids these issues and outperforms single-variable models across all tested variables. These results highlight the effectiveness of our multi-variable architecture for high-resolution climate downscaling.

\subsection{Data Selection}\label{sec:Data}

Our study is based on the EURO-CORDEX data. EURO-CORDEX, developed as part of the Coordinated Regional Downscaling Experiment (CORDEX) under the World Climate Research Programme (WCRP), provides a consistent set of high-resolution (0.11°) RCM simulations, driven by boundary conditions from multiple GCMs and emission scenarios. These simulations are widely used for regional climate impact assessments across Europe \cite{Jacob2014}. We subjectively selected one GCM-RCM couple from the available simulations: the MPI-ESM-LR GCM with the ICTP-RegCM4-6 RCM simulation under emissions scenario RCP8.5. 

The selection of surface air temperature (tas), surface wind speed (sfcWind), geopotential height at 500 hPa (zg500), total precipitation (tp), surface downwelling shortwave radiation (rsds), and surface downwelling longwave radiation (rlds) was motivated by both practical relevance and methodological interest. These six variables exhibit markedly different spatial and statistical characteristics when downscaled. Temperature and geopotential height produce relatively smooth, large-scale fields, whereas wind speed contains localized high-frequency patterns. Precipitation, in contrast, is sparse, intermittent, and notoriously challenging to downscale. Radiative fluxes (rsds and rlds) reflect a mixture of smooth background structure and localized effects driven by cloud cover. This diversity offers a useful testbed for evaluating how well data-driven models handle different signal types, and whether modeling them jointly can facilitate positive transfer across variables.

The choice was also shaped by the broader goal of exploring multi-variable downscaling. By selecting a diverse set of variables, we can begin to assess whether downscaling performance benefits from shared representations across related but physically distinct fields. The downscaling task addresses a spatial resolution increase from 1.0° to 0.1°, representing a tenfold enhancement. This is among the largest resolution gaps available in publicly accessible GCM-RCM datasets, offering a particularly demanding test case for evaluating the capabilities of data-driven downscaling methods.

The domain of interest is Europe, selected for its climatic diversity, data availability, and relevance to policy and research communities. To ensure a direct comparison between scales, the low-resolution GCM data was interpolated onto the RCM’s 0.1° grid using bilinear interpolation. This pre-processing step ensures that GCM and RCM fields are directly comparable in both space and time. The datasets span daily time steps from 2006 to 2099.

\subsection{Related Works}\label{sec:Related_Works}
In recent years, the focus of climate downscaling has increasingly shifted toward deep learning models \cite{sun2024deep, bano2021suitability, bano2020configuration}. A wide range of architectures has been explored to downscale variables such as temperature, wind components, surface pressure, and precipitation, among others. These include convolutional neural networks (CNNs) \cite{bano2021suitability, wang2021deep, dujardin2022wind, jin2022deep, damiani2024exploring, vandal2017deepsd}, U-Nets \cite{sha2020deep, hohlein2020comparative, doury2023regional}, Vision Transformers (ViTs) \cite{zhong2024investigating, yang2024successful, liu2025spatial}, generative adversarial networks (GANs) \cite{harris2022generative, kumar2023modern, stengel2020adversarial, manco2025new}, and diffusion models \cite{aich2024conditional, mardani2023generative, merizzi2024wind, merizzi2025controlling, watt2024generative}.

Different architectures show varying strengths depending on the target variable. Generative models like GANs and diffusion models are particularly effective at preserving high-frequency spatial detail and have shown strong performance in downscaling wind speed \cite{tomasi2025can}. In contrast, smoother fields such as temperature are often better handled by U-Nets \cite{liu2024enhanced}. Despite these findings, there is no clear consensus on a universally optimal architecture. Recently, ViT-based models have shown competitive and often superior performance across multiple climate variables, outperforming CNN-based baselines in several studies \cite{loganathan2025regional, chen2022rainnet, zhong2024investigating}. As such, ViTs form the foundation of our proposed approach.

ViT-based architectures for image-to-image tasks often draw inspiration from the encoder-decoder structure of U-Nets. One common design integrates a transformer-based encoder with a convolutional decoder \cite{chen2021transunet, ranftl2021vision, liang2021swinir}, while other variants opt for fully transformer-based pipelines \cite{wang2022uformer, zheng2022ittr}. A key architectural consideration is the treatment of skip connections, which preserve spatial resolution and improve gradient flow. These may be preserved directly \cite{chen2021transunet, zhang2025alternate}, omitted entirely under the assumption of rich encoder representations \cite{ranftl2021vision}, or implemented through attention-based gating mechanisms \cite{wang2022uformer}.

While the majority of deep learning-based downscaling studies employ single-variable models, where each variable is modeled independently, there is growing interest in multi-variable approaches that leverage cross-variable relationships. These approaches aim to enhance physical consistency across variables and potentially improve accuracy. In this work, we design a multi-variable downscaling framework and benchmark its performance against single-variable counterparts.

Multi-variable downscaling naturally aligns with the framework of multi-task learning, where a single model learns to solve multiple related tasks simultaneously. In climate applications, multi-task learning offers several potential advantages: improved generalization through shared representations, reduced training and inference cost, and the possibility of positive transfer across variables \cite{crawshaw2020multi}.

Several recent studies support the promise of this approach \cite{schillinger2025enscale}. For instance, Ling et al. \cite{ling2022multi} present a multi-task model for seasonal climate prediction that jointly forecasts several indices related to the Indian Ocean Dipole. The model is trained using a loss function defined as a weighted sum of the individual task with a higher loss weight to the primary target (DMI). The model achieves improved performance compared to single-variable baselines. Similarly, Fan et al. \cite{Fan2020} propose M2GSNet, a graph-based multi-task model for wind power prediction across multiple farms. By utilizing shared encoders across tasks and training with an unweighted sum of L1 losses across farms, the model achieves superior forecasting accuracy compared to single-variable baselines.

Other examples further support the benefits of multi-task formulations. Bannai et al. \cite{bannai2023multi} use a shared CNN backbone with two heads for simultaneous rain/no-rain classification and rain rate estimation, yielding improved performance through joint optimization of a combined loss function. Liu et al. \cite{liu2023dsaf} propose a two-stage downscaling system incorporating bias correction, terrain inputs, and a multi-task loss with physical constraints, demonstrating the potential of multi-variable models for enhancing both accuracy and realism in climate data products. Quesada-Chacón et al. \cite{quesada2023downscaling} propose a deep learning-based downscaling framework that employs U-Net architectures to generate daily 1km resolution gridded ensembles of seven meteorological variables, including precipitation, temperature (min/mean/max), radiation, humidity, and wind.

\section{Methodology}

\begin{figure}[h]
\centering
\includegraphics[width=\linewidth]{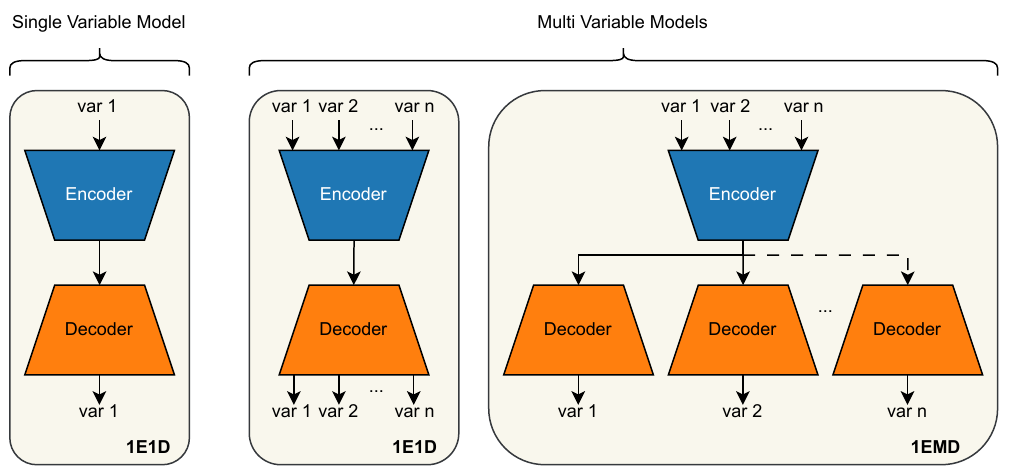}
\caption{Comparison between a single-variable encoder-decoder model and multi-variable models using either a single shared decoder (1E1D) or multiple decoders branching from a shared encoder (1EMD).}
\label{fig:different_model_abstract}
\end{figure}

In this section, we start by detailing our backbone implementation of a ViT-based image-to-image model. We then proceed to describe the single-variable approach, where separate encoder-decoder architectures are trained independently for each target variable. Next, introduce two multi-variable strategies. The first, a single-encoder single-decoder (1E1D) model, concatenates multiple variables along the channel dimension, treating them as multi-channel images in a unified ViT pipeline, and the second a single-encoder multi-decoder (1EMD) where a shared encoder extracts common features, while separate decoders are dedicated to reconstructing each variable individually. A visual comparison of the proposed approaches is shown in Figure \ref{fig:different_model_abstract}.

\subsection{Architectural Backbone: Vision Transformer (ViT)}\label{Section:VIT}

Vision Transformers (ViTs) are neural architectures that originate in natural language processing, where self-attention is used to capture relationships within sequences. In the vision domain, ViT applies the same principle by splitting an image into fixed-size patches, which are then flattened, embedded, and treated as a sequence. This sequence is processed by transformer blocks, allowing the model to capture global image dependencies without relying on convolutions, effectively serving as an alternative to traditional encoder networks.

In recent years ViTs have been successfully applied to a variety of super-resolution tasks \cite{lu2022transformer, 10836746, cao2021video, liu2023image}. Our proposed ViT architecture belongs to the ViT-CNN hybrid category, combining a transformer-based encoder with a convolutional decoder, enabling the model to capture global, non-local relationships more effectively than purely convolutional networks \cite{chen2021transunet, liang2021swinir}. This design allows the model to handle complex spatial patterns, which is especially important for tasks that require modeling long-range and cross-variable dependencies.

Skip connections are a well-established component of encoder-decoder architectures, commonly used to preserve spatial detail and improve gradient flow. In ViT-CNN hybrid models, studies have shown that skip connections can enhance performance by bridging lower-level features from the encoder to the decoder \cite{chen2021transunet}. However, in transformer-based architectures,especially those without hierarchical designs, this benefit may be less critical \cite{ranftl2021vision, zheng2021rethinking}. Unlike CNN encoders, which typically reduce spatial resolution through downsampling, our ViT encoder preserves a constant resolution throughout. As a result, the output tokens retain the same spatial structure as the input, reducing the need for skip connections to recover lost detail.

In our implementation, we chose to omit skip connections to prioritize modularity and simplicity, particularly in the context of supporting multiple variable-specific decoders. In a multi-decoder setting, implementing skip connections would require duplicating and routing encoder features to each decoder, increasing architectural complexity and coupling. Moreover, since each decoder specializes in a different output variable, sharing low-level features indiscriminately across all decoders could introduce unintended interactions or noise between tasks, potentially harming performance. Given that the empirical benefits of skip connections in transformer encoders remain inconclusive, we deemed the flexibility of a skip-free design—enabling easy addition or removal of decoders—more valuable than a potentially marginal performance gain.

Specifically, our architecture is designed as follows: The encoding path follows a standard ViT configuration, where the input image is divided into non-overlapping $8 \times 8$ patches, each linearly projected into a 256-dimensional vector by a dense embedding layer. To retain spatial information, learnable positional embeddings are added to the patch representations before passing them through the transformer stack. The encoder consists of six transformer layers, each following a consistent structure. Each layer applies normalization, followed by multi-head self-attention with six heads and a dropout rate of 0.1. Next, A residual connection adds the input of the attention block to its output. After attention, a second normalization is applied, followed by a multi-layer perceptron (MLP) consisting of two fully connected layers with 512 and 256 units, respectively, and GELU \cite{hendrycks2016gaussian} activation. Another residual connection is applied at the output of the MLP block. Through this process, the transformer encoder learns both local and global relationships between spatially distributed patches. To mitigate the memory constraints that limit batch size, we use Group Normalization \cite{Wu2018} with a group size of 8, which has been shown to be more stable than Batch Normalization in such settings.

Following the transformer encoder, the sequence of embeddings is reshaped into a spatial grid, initializing the decoding path. The decoder is composed of three consecutive upsampling blocks, each of which includes bilinear interpolation for upsampling, doubling the spatial resolution at each step. We utilize bilinear interpolation avoiding artifacts often associated with learned transposed convolutions \cite{odena2016deconvolution}. Each bilinear upsampling is followed by a residual block, composed of group normalization, two convolutional layers, and a residual connection that preserves input information. The first convolution in the block uses Swish activation, while the second uses no activation. The feature width begins at 256 in the first upsampling block and is halved in each subsequent block, reducing to 32 in the final block. A final convolutional layer with no activation and a kernel initializer set to zero projects the output to the target resolution, completing the decoding process. The complete architecture for a single encoder-decoder pair contains approximately 11 million learnable parameters.

\subsection{Single-Variable Models}\label{sec:single_var_models}

The most direct approach at downscaling multiple variables is based on the use of multiple single-variable models trained independently on each variable, with no sharing of weights or other information between each model. 

Each individual model consists of a single encoder-decoder pair, which enables the model to focus on the individual task but doesn't enable the utilization of cross-variable information. 

In the case of single variable the loss is calculated on the comparison between the generated output and the available ground truth, with our implementation using mean square error (MSE) loss.  Denoting $\hat{Y}{hw}$ as the predicted pixel at spatial position $(h, w)$ and $Y{hw}$ as the corresponding ground truth pixel, the loss is defined as:

\begin{equation}
    \mathcal{L}_{\text{MSE}} = \frac{1}{HW} \sum_{h=1}^{H} \sum_{w=1}^{W} \left( \hat{Y}_{hw} - Y_{hw} \right)^2
\end{equation}

\subsection{Multi-Variable Models}\label{sec:multi_var_models}

The main goal of this work is to design an architecture capable of downscaling multiple variables simultaneously. While the coarse-resolution GCM provides the primary input, we hypothesize that additional value can be gained by exploiting inter-variable relationships. Our assumption is that the atmospheric state encodes joint spatial and dynamical patterns across variables, which a shared representation can capture more effectively than isolated models. By modeling all variables together, the network can leverage these patterns, potentially leading to improved accuracy and greater physical consistency compared to independent downscaling approaches.

Building upon the ViT architecture described in Section~\ref{Section:VIT}, we implement two multi-variable variants: a Single-Encoder Single-Decoder (1E1D) model, where all variables share the same encoder-decoder pipeline, and a Single-Encoder Multi-Decoder (1EMD) model, where the variable are jointly encoded but decoded separately. These designs allows us to study the trade-off between fully shared architectures and variable-specific specialization via independent decoders.

Both our proposed multi-variable approaches share the same loss structure.  Training is performed simultaneously across the six target variables, with the loss computed as the mean over all output channels. To make this approach viable, it is essential to normalize all variables to the same range. This ensures that the model optimizes all variables with equal priority, preventing any one variable from dominating the loss due to scale differences. With MSE being our loss function of choice, and denoting $\hat{Y}{nhw}$ as the predicted pixel at spatial position $(h, w)$ for variable $n$, and $Y{nhw}$ as the corresponding ground truth pixel, the joint loss is defined as:

\begin{align}
\mathcal{L}_{\text{MSE}} = \frac{1}{NHW} \sum_{n=1}^{N} \sum_{h=1}^{H} \sum_{w=1}^{W} 
&\left( \hat{Y}_{nhw} - Y_{nhw} \right)^2 
\end{align}

\subsubsection{Single-Encoder Single-Decoder VIT (1E1D)}

Our first approach to multi-variable modeling treats the different atmospheric fields as complementary features of a single spatial input by concatenating the variables along the channel dimension. This design keeps the single-variable architecture largely intact, with modifications only to the patch embedding and final convolutional layers to accommodate six input and output channels.

The input consists of six atmospheric variables stacked as a six-channel image. We divide this input into non-overlapping patches of shape $8 \times 8 \times 6$, each of which is projected into a 256-dimensional vector via a dense embedding layer. These token embeddings are then processed by a Vision Transformer (ViT) encoder, following the same architecture used in the single-variable model. The decoder, composed of convolutional layers, upsamples the latent representation back to the original spatial resolution, with the final layer producing six output channels corresponding to the predicted fields.

This approach allows the encoder to capture a joint representation of the atmospheric state across multiple variables. However, it implicitly assumes that the variables are spatially aligned and compatible when treated as separate channels, an assumption that may be limiting if the variables exhibit different spatial structures or temporal dynamics.

\subsubsection{Single-Encoder Multi-Decoder VIT (1EMD)}

The six target variables in our downscaling task exhibit substantial differences in their spatial structures and statistical distributions. For example, temperature fields tend to show smoother gradients compared to the more intricate patterns found in wind components. When using a single decoder to reconstruct all variables jointly, these differences can cause interference: the decoder is limited to sharing the same weights for generating the different variables, leading to potential cross-variable contamination. In particular, we observed that high-frequency patterns from one variable (e.g., wind) can bleed into the smoother reconstructions of another (e.g., geopotential height, tp), degrading performance.

To address this, we propose a novel multi-variable downscaling architecture that employs a single transformer-based encoder shared across all variables, followed by a dedicated convolutional decoder for each output variable. This design preserves the benefit of jointly encoding spatial inter-variable dependencies while allowing each decoder to specialize in reconstructing a specific variable. By decoupling the decoding paths, we reduce interference and provide sufficient representational capacity to each output stream. 
We consider this architecture to be an effective compromise between the two aforementioned approaches: running completely independent encoder-decoder pipelines for each variable, and forcing all variables through a single encoder-decoder with channel concatenation. Our approach leverages shared context while controlling model complexity and avoiding cross-variable entanglement, achieving a favorable trade-off between accuracy and computational cost.

\begin{figure}[h!]
\centering
\includegraphics[width=0.99\linewidth]{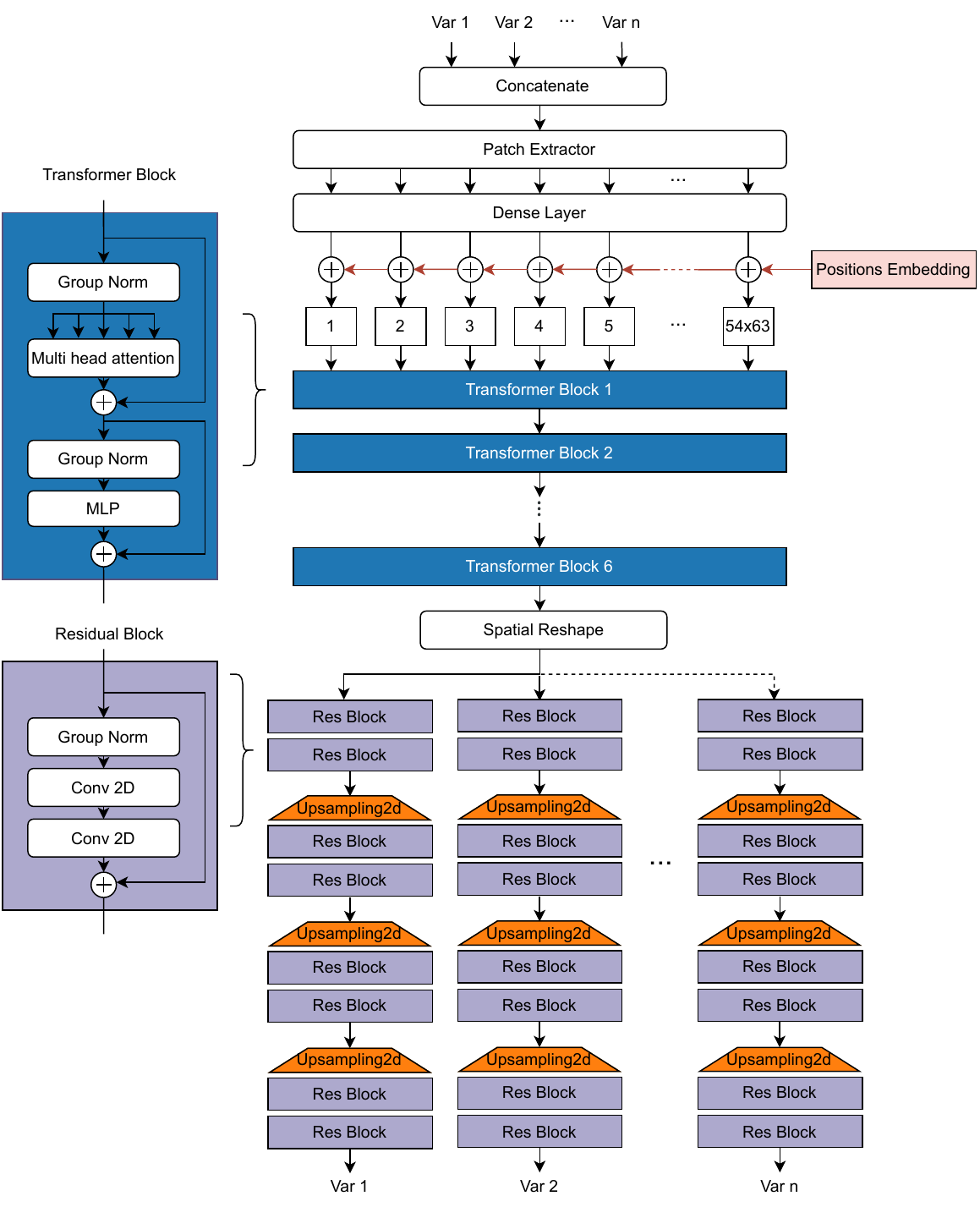}
\caption{Architecture of the multi-variable single-encoder multi-decoder model (1EMD). Concatenated input variables are divided into patches, embedded with positional encodings, and processed as a sequence by six transformer blocks,  each consisting of multi-head self-attention (6 heads) followed by an MLP. The output sequence is then reshaped into a spatial grid and passed to separate CNN-based decoders, each composed of three upsampling stages interleaved with residual blocks.}
\label{fig:multi-variable-model}
\end{figure}

Specifically, the 1EMD architecture is structured as follows. Each low-resolution input variable is first bilinearly upsampled to the target resolution of $504 \times 432$, matching the output grid. The resulting variables are concatenated along the channel dimension and partitioned into 3402 non-overlapping patches of size $8 \times 8 \times 6$. Each patch is linearly embedded into a 256-dimensional vector and augmented with positional information via additive encoding. The sequence of embedded patches is processed by a stack of six transformer blocks, each featuring multi-head self-attention (six heads), residual connections with group normalization, and MLP layers with GELU activations.

The final encoder output is reshaped into a spatial feature grid of size $54 \times 63 \times 256$ and passed in parallel to six independent convolutional decoders, one per variable. The decoders do not share parameters and are trained jointly using a composite loss function that accounts for all variables. Each decoder consists of a sequence of residual blocks with group normalization and Swish activations, interleaved with 2D upsampling layers. Three upsampling stages are used to restore the original resolution of $504 \times 432$. The final output is produced by a zero-initialized convolutional layer without activation.
The complete 1EMD model architecture is reported in Figure \ref{fig:multi-variable-model}. 

Experimental results, discussed in detail in Section \ref{sec:results}, demonstrate that the proposed 1EMD model outperforms both the single-variable and 1E1D approaches.

\subsubsection{Multi-Variable U-net}

In imaging tasks, the U-Net architecture represents a widely adopted standard baseline, with applications spanning much of the image-to-image translation spectrum. U-Nets have also been extensively used for statistical downscaling of meteorological variables, including several recent contributions specifically addressing multivariate downscaling \cite{alipourhajiagha2025probabilistic, quesada2023downscaling, li2025generative}.
To provide a meaningful and broadly recognized comparison against our transformer-based single-variable and multi-variable approaches, we include a multi-variable U-Net structured in a 1E1D configuration.

The implemented residual U-Net follows a standard symmetric encoder–decoder design built around residual blocks. Each residual block applies group normalization followed by two 3×3 convolutional layers. The encoder is composed of three downsampling stages, each containing three residual blocks and a 2× spatial reduction achieved via average pooling. Feature maps from every block are stored as skip connections and later concatenated in the decoder. The bottleneck stage processes the lowest-resolution representation through multiple residual blocks at the widest channel depth. The decoder mirrors the encoder, performing 2× bilinear upsampling at each stage and concatenating the corresponding skip features, followed by residual blocks

In the multi-variable, multi-task setting, the six variables are concatenated along the channel dimension in the input. A symmetric multi-channel output head is used to reconstruct the six variables, preserving a direct correspondence between input and output channels. The sizing of the model was designed to match the competing ViT based architectures in terms of execution and training times, The final configuration comprises approximately five million trainable parameters.

\section{Experimental Setup}
Our objective is to learn a mapping from coarse-resolution Global Climate Model (GCM)  to high-resolution Regional Climate Model (RCM). Specifically, we use data from the MPI-ESM-LR GCM and the ICTP-RegCM4-6 RCM under the RCP8.5 scenario, addressing a spatial resolution shift from $1\degree$ (approximately 100 km) to 0.1\degree (approximately 10 km). The GCM data are re-gridded to the RCM domain using bilinear interpolation to match the spatial structure of the RCM over Europe.

We focus on downscaling six key atmospheric variables: near-surface air temperature (tas), surface wind speed (sfcWind), 500 hPa geopotential height (zg500), total precipitation (tp), surface downwelling shortwave radiation (rsds), and surface downwelling longwave radiation (rlds). From an image processing perspective, each variable is treated as a 2D field with spatial dimensions of $431\times501$ pixels.

Our dataset spans daily values from 2006 to 2100. We use data from 2006--2099 for training and reserve one year per decade (2010, 2020, etc.) for testing. This design avoids long-term temporal separation between training and test sets, helping prevent bias from trends such as climate change. For each variable, GCM and RCM outputs are temporally and spatially aligned so that the models can effectively learn to predict high-resolution RCM fields from corresponding low-resolution GCM inputs.

As of pre-processing, each variable is min-max normalized to the [0, 1] range using the training data's minimum and maximum values. To ensure compatibility with convolutional architectures, images are also padded via replication to $432\times504$ pixels to make the dimensions divisible by 8. Padding is removed before computing evaluation metrics.

Our downscaling models operate in an image-to-image manner, taking the GCM variable(s) at a given timestamp as input and predicting the corresponding RCM variable(s). In the single-variable case, the model maps one GCM field to one RCM field. In the multi-variable case, the model maps all six GCM variables to their corresponding RCM fields for the same timestamp, effectively forming a six-to-six mapping as illustrated in Figure \ref{fig:different_model_abstract}.

We adopt a fair comparison framework that is coherent with standard practice in model comparison, avoiding common pitfalls such as unequal fine-tuning, training effort or model complexity \cite{lucic2018gans, pineau2021improving, lipton2019troubling}. We compare the four modeling approaches, three based on a Vision Transformer (ViT) backbone and one based on a U-Net. First, we train and test six independent single-variable ViT models, one for each target variable. Then, we explore three multi-variable models: 1E1D, which uses a shared encoder and decoder, 1EMD, which employs a shared encoder and separate decoders for each variable, and a multi-variable U-Net providing an external, widely used reference architecture.
All ViT models share the same encoder architecture. The single-variable ViT and the multi-variable ViT 1E1D also share the same parameter count, differing only in the final convolutional layer and decoder embeddings. In contrast, the 1EMD model maintains a shared encoder but uses six independent decoders (one per variable), with no weight sharing in the decoding stage.

We use the Adam optimizer with decoupled weight decay (AdamW) \cite{loshchilov2017decoupled}, a common choice for ViT-based architectures. All models are trained for 400 epochs with 500 steps per epoch. Due to memory constraints, we use a batch size of 1. The learning rate is fixed at $1\mathrm{e}{-4}$ for the first 300 epochs and reduced to $1\mathrm{e}{-5}$ for the final 100 epochs. Weight decay is set to one-tenth of the current learning rate. These settings were chosen based on empirical tuning to ensure consistent convergence across all approaches. We use mean squared error (MSE) as the loss function for all models, as described in Sections~\ref{sec:single_var_models} and~\ref{sec:multi_var_models}.

Training is conducted on a NVIDIA A100 GPU using a unified pre-processing pipeline. Testing is performed on the reserved ten years of test data, spanning the full 100-year period. Each model is evaluated on the entire test set, and performance metrics are computed over all test samples. 

To account for variability due to random initialization and stochastic optimization, all models were trained and evaluated twice using independent random seeds. Each repetition executed the full training and testing pipeline under identical settings, and the performance values reported in the manuscript correspond to the mean across the two runs. This protocol provides a lightweight but meaningful assessment of robustness within realistic computational constraints. Notably, the relative ranking of the models is consistent across both repetitions.

\section{Results}\label{sec:results}

% \begin{figure}[h]
% \centering
% \includegraphics[width=\linewidth]{361.png}
% \caption{Comparison between all the investigated models on a single sample from the test set}
% \label{fig:multi-variable-model}
% \end{figure}
% \begin{figure}[h!]
% \centering

% \begin{subfigure}{\linewidth}
%     \centering
%     \includegraphics[width=\linewidth]{361.png}
%     \caption{}
% \end{subfigure}
% \begin{subfigure}{\linewidth}
%     \centering
%     \includegraphics[width=\linewidth]{763.png}
%     \caption{}
% \end{subfigure}
% % \begin{subfigure}{\linewidth}
% %     \centering
% %     \includegraphics[width=\linewidth]{2596.png}
% % \end{subfigure}

% \caption{Comparison of the examined models (Single var ViT, multi var ViT 1E1D, Multi var ViT 1EMD), target RCM,  and bilinearly interpolated GCM for two different samples of the test set. For each sample, each row contains geopotential, wind speed and temperature for the same timestamp.}
% \label{fig:model-comparison}
% \end{figure}

\begin{figure}[h!]
\centering
\includegraphics[width=\linewidth]{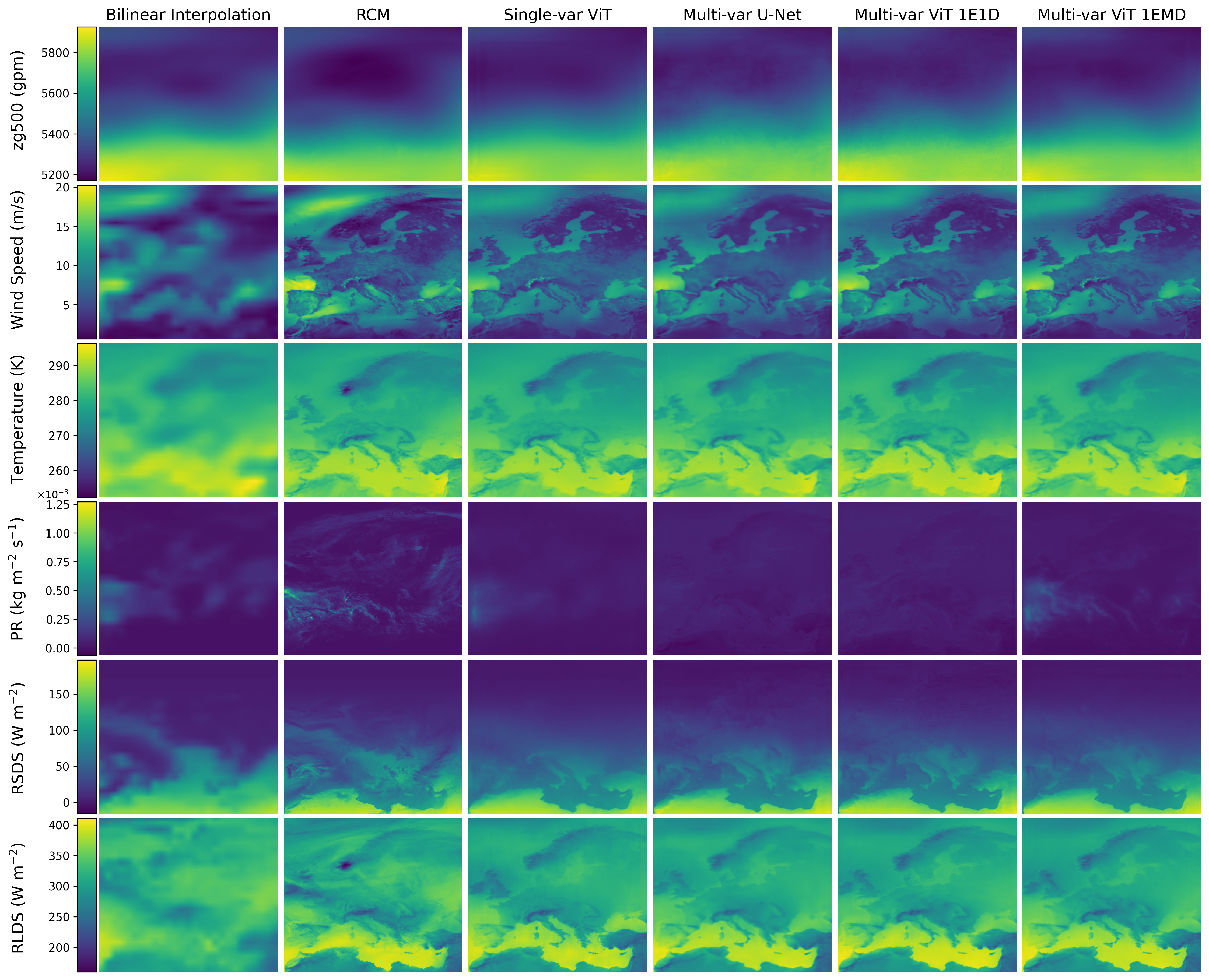}
\caption{Comparison of the examined models (Single var ViT, multi var ViT 1E1D, Multi var ViT 1EMD), target RCM,  and bilinearly interpolated GCM for a sample of the test set. For each sample, each row contains Geopotential, Wind speed, Temperature, Total Precipitation, Surface Downwelling Shortwave Radiation and Surface Downwelling Longwave Radiation for the same timestamp.}
\label{fig:model-comparison}
\end{figure}

\begin{table}[htbp]
\centering
\def\arraystretch{0.9}
\resizebox{\textwidth}{!}{%
\begin{tabular}{>{\bfseries}p{5cm}ccc}
\toprule
\multicolumn{4}{c}{\textbf{Mean Performance Across Two Independent Runs}} \\
\midrule
\textbf{Model} & \textbf{MSE $\downarrow$} & \textbf{MAE $\downarrow$} & \textbf{SSIM $\uparrow$} \\
\midrule
\multicolumn{4}{c}{\textbf{tas}} \\
Bilinear                               & 9.97e-04 & 2.27e-02 & 9.78e-01 \\
Single Var. ViT                        & 4.34e-04 & 1.47e-02 & \underline{9.89e-01} \\
Multi Var. ViT (1E1D)                  & 4.50e-04 & 1.51e-02 & 9.84e-01 \\
Multi Var. U-Net                       & 4.47e-04 & 1.48e-02 & 9.83e-01 \\
Multi Var. ViT (1EMD)                  & \underline{4.23e-04} & \underline{1.43e-02} & 9.88e-01 \\
\midrule
\multicolumn{4}{c}{\textbf{sfcWind}} \\
Bilinear                               & 6.08e-03 & 5.81e-02 & 8.35e-01 \\
Single Var. ViT                        & 3.43e-03 & 4.36e-02 & 8.85e-01 \\
Multi Var. ViT (1E1D)                  & 3.33e-03 & 4.29e-02 & 8.76e-01 \\
Multi Var. U-Net                       & 3.35e-03 & 4.32e-02 & 8.69e-01 \\
Multi Var. ViT (1EMD)                  & \underline{3.23e-03} & \underline{4.21e-02} & \underline{8.87e-01} \\
\midrule
\multicolumn{4}{c}{\textbf{zg500}} \\
Bilinear                               & 2.55e-03 & 3.76e-02 & \underline{9.95e-01} \\
Single Var. ViT                        & 1.90e-03 & 3.18e-02 & \underline{9.95e-01} \\
Multi Var. ViT (1E1D)                  & 1.85e-03 & 3.14e-02 & 9.94e-01 \\
Multi Var. U-Net                       & 1.84e-03 & 3.17e-02 & \underline{9.95e-01} \\
Multi Var. ViT (1EMD)                  & \underline{1.80e-03} & \underline{3.08e-02} & \underline{9.95e-01} \\
\midrule
\multicolumn{4}{c}{\textbf{tp}} \\
Bilinear                               & 6.00e-07 & 3.73e-04 & \underline{9.97e-01} \\
Single Var. ViT                        & 4.46e-07 & 3.55e-04 & \underline{9.97e-01} \\
Multi Var. ViT (1E1D)                  & 4.91e-07 & 3.94e-04 & 9.96e-01 \\
Multi Var. U-Net                       & 5.00e-07 & 4.09e-04 & 9.96e-01 \\
Multi Var. ViT (1EMD)                  & \underline{4.32e-07} & \underline{3.48e-04} & \underline{9.97e-01} \\
\midrule
\multicolumn{4}{c}{\textbf{rsds}} \\
Bilinear                               & 2.24e-02 & 1.01e-01 & 7.79e-01 \\
Single Var. ViT                        & 5.37e-03 & 5.02e-02 & 8.80e-01 \\
Multi Var. ViT (1E1D)                  & 5.03e-03 & 4.86e-02 & 8.76e-01 \\
Multi Var. U-Net                       & 5.00e-03 & 4.85e-02 & 8.75e-01 \\
Multi Var. ViT (1EMD)                  & \underline{4.99e-03} & \underline{4.83e-02} & \underline{8.83e-01} \\
\midrule
\multicolumn{4}{c}{\textbf{rlds}} \\
Bilinear                               & 6.70e-03 & 6.37e-02 & 9.17e-01 \\
Single Var. ViT                        & 3.06e-03 & 4.28e-02 & 9.43e-01 \\
Multi Var. ViT (1E1D)                  & 2.80e-03 & 4.03e-02 & 9.43e-01 \\
Multi Var. U-Net                       & 2.81e-03 & 4.05e-02 & 9.40e-01 \\
Multi Var. ViT (1EMD)                  & \underline{2.78e-03} & \underline{4.01e-02} & \underline{9.47e-01} \\
\bottomrule
\end{tabular}
}
\caption{Mean model performance across two independent runs. Best values (including ties) are underlined.}
\label{tab:variable_results_exec}
\end{table}

We evaluate the four models, Single-variable ViT, Multi-variable ViT (1E1D), Multi-variable ViT (1EMD) and Multi-variable U-Net on a test dataset comprising ten years of unseen data. The test set includes one year per decade from 2010 to 2100.

All models are tested across the six selected variables using three evaluation metrics: Mean Squared Error (MSE), Mean Absolute Error (MAE), and Structural Similarity Index (SSIM) alongside spatial and temporal Pearson correlations. These metrics offer a well-rounded perspective, balancing numerical accuracy (MSE, MAE) with perceptual similarity (SSIM), and are commonly adopted in climate imaging studies \cite{bano2020configuration,asperti2025precipitation, yang2024fourier, harder2023hard, oyama2023deep}. All metrics are computed on normalized data to enable cross-variable comparison as the variables differ in their units and ranges.

\begin{figure}[h!]
\centering
\includegraphics[width=\linewidth]{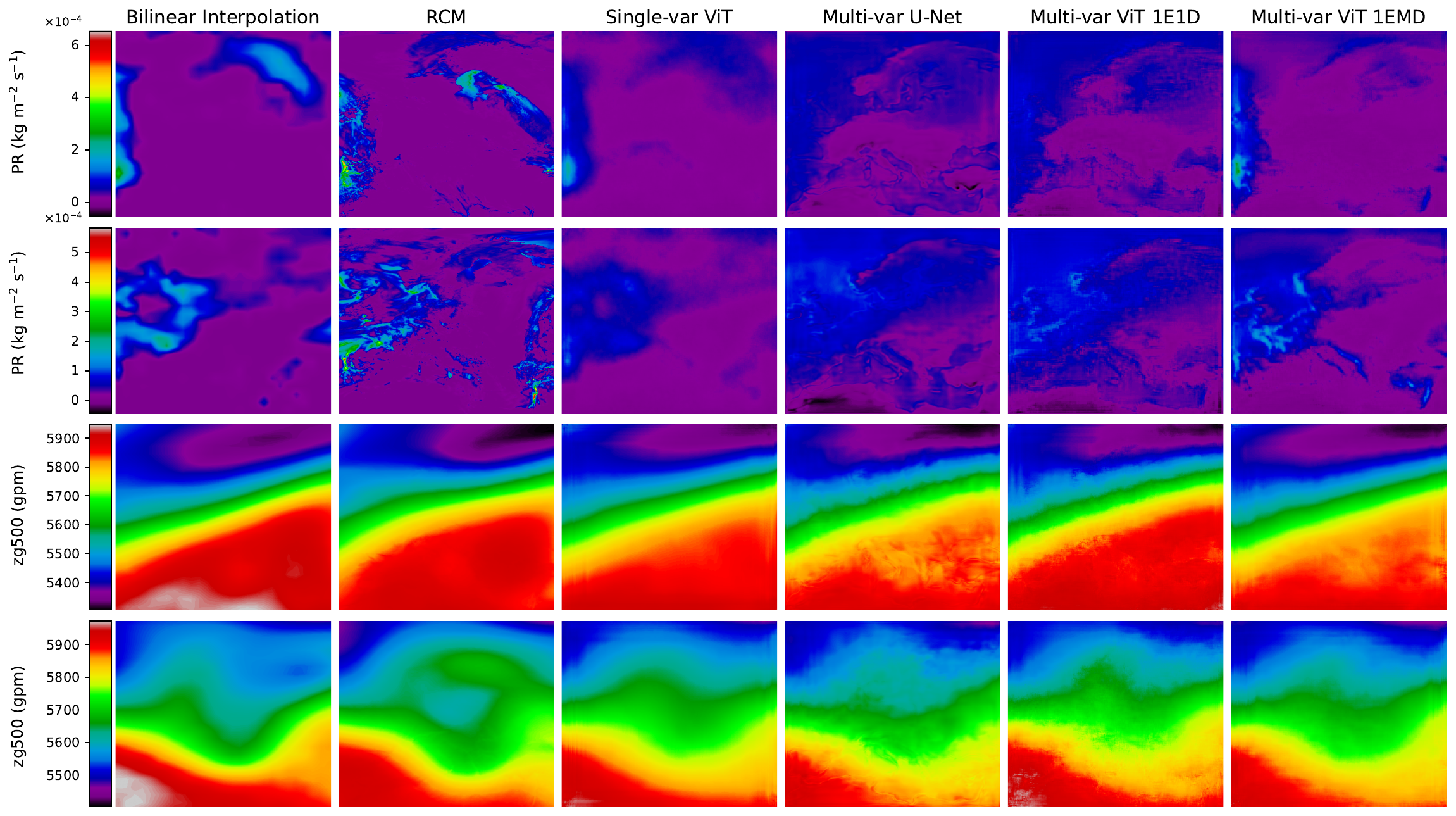}
\caption{Visualization of cross-variable leakage in the 1E1D and multi-variable U-Net models for total precipitation and zg500. Four samples are shown, two for precipitation and two for geopotential height. Using a high contrast colormap, we can clearly see high-frequency artefacts in the multi-variable U-Net and 1E1D ViT, with faint outlines resembling European geography. This suggests contamination from temperature and wind fields. These outlines are absent from the original data, the single-variable model, and the multi-decoder 1EMD model.}
\label{fig:cross_variable_artefacts}
\end{figure}

\begin{table}[htbp]
\centering
\def\arraystretch{1}
\resizebox{\textwidth}{!}{%
\begin{tabular}{>{\bfseries}p{4cm}cc}
\toprule
\multicolumn{3}{c}{\textbf{Spatial and Temporal Correlations (MPI to RegCM)}} \\
\midrule
\textbf{Model} & \textbf{Spatial corr $(\mu \pm \sigma)$} & \textbf{Temporal corr $(\mu \pm \sigma)$} \\
\midrule
\multicolumn{3}{c}{\textbf{tas}} \\
Bilinear              & $0.910 \pm 0.037$ & $0.935 \pm 0.016$ \\
Single Var. ViT       & $0.957 \pm 0.024$ & $0.956 \pm 0.011$ \\
Multi Var. ViT (1E1D) & $0.957 \pm 0.024$ & $0.956 \pm 0.010$ \\
Multi Var. U-Net      & $0.955 \pm 0.024$ & $0.956 \pm 0.011$ \\
Multi Var. ViT (1EMD) & $\underline{0.958 \pm 0.024}$ & $\underline{0.957 \pm 0.010}$ \\
\midrule
\multicolumn{3}{c}{\textbf{sfcWind}} \\
Bilinear              & $0.602 \pm 0.161$ & $0.528 \pm 0.096$ \\
Single Var. ViT       & $0.748 \pm 0.113$ & $0.605 \pm 0.083$ \\
Multi Var. ViT (1E1D) & $0.759 \pm 0.109$ & $0.636 \pm 0.075$ \\
Multi Var. U-Net      & $0.755 \pm 0.111$ & $0.631 \pm 0.081$ \\
Multi Var. ViT (1EMD) & $\underline{0.765 \pm 0.109}$ & $\underline{0.644 \pm 0.073}$ \\
\midrule
\multicolumn{3}{c}{\textbf{zg500}} \\
Bilinear              & $0.929 \pm 0.089$ & $0.920 \pm 0.033$ \\
Single Var. ViT       & $0.943 \pm 0.073$ & $0.935 \pm 0.024$ \\
Multi Var. ViT (1E1D) & $0.943 \pm 0.070$ & $0.937 \pm 0.023$ \\
Multi Var. U-Net      & $\underline{0.944 \pm 0.070}$ & $\underline{0.938 \pm 0.023}$ \\
Multi Var. ViT (1EMD) & $\underline{0.944 \pm 0.070}$ & $0.937 \pm 0.023$ \\
\midrule
\multicolumn{3}{c}{\textbf{tp}} \\
Bilinear              & $0.339 \pm 0.173$ & $0.340 \pm 0.111$ \\
Single Var. ViT       & $0.411 \pm 0.158$ & $0.389 \pm 0.112$ \\
Multi Var. ViT (1E1D) & $0.341 \pm 0.121$ & $0.321 \pm 0.116$ \\
Multi Var. U-Net      & $0.293 \pm 0.142$ & $0.298 \pm 0.128$ \\
Multi Var. ViT (1EMD) & $\underline{0.438 \pm 0.150}$ & $\underline{0.407 \pm 0.106}$ \\
\midrule
\multicolumn{3}{c}{\textbf{rsds}} \\
Bilinear              & $0.745 \pm 0.165$ & $0.826 \pm 0.048$ \\
Single Var. ViT       & $0.885 \pm 0.093$ & $0.946 \pm 0.016$ \\
Multi Var. ViT (1E1D) & $\underline{0.894 \pm 0.088}$ & $\underline{0.951 \pm 0.014}$ \\
Multi Var. U-Net      & $0.893 \pm 0.088$ & $0.950 \pm 0.013$ \\
Multi Var. ViT (1EMD) & $\underline{0.894 \pm 0.087}$ & $\underline{0.951 \pm 0.014}$ \\
\midrule
\multicolumn{3}{c}{\textbf{rlds}} \\
Bilinear              & $0.486 \pm 0.168$ & $0.703 \pm 0.078$ \\
Single Var. ViT       & $0.791 \pm 0.089$ & $0.811 \pm 0.053$ \\
Multi Var. ViT (1E1D) & $\underline{0.808 \pm 0.086}$ & $\underline{0.831 \pm 0.054}$ \\
Multi Var. U-Net      & $0.805 \pm 0.088$ & $0.830 \pm 0.055$ \\
Multi Var. ViT (1EMD) & $\underline{0.808 \pm 0.088}$ & $0.829 \pm 0.054$ \\
\bottomrule
\end{tabular}
}
\caption{Spatial and temporal Pearson correlations between model predictions and RegCM4-6 targets for each variable. 
Spatial correlations are computed for each day and summarized as mean $\pm$ standard deviation over time. 
Temporal correlations are computed at each grid point and summarized as mean $\pm$ standard deviation over space. 
Best values (including ties) are underlined.}
\label{tab:spatiotemporal_corr}
\end{table}

% \begin{figure}[h!]
% \centering
% \includegraphics[width=\linewidth]{errors_6variables.png}
% \caption{Mean absolute error maps comparing the performance of bilinear interpolation (top row) and the multi-variable, multi-decoder 1EMD ViT model (bottom row). Each column shows the spatial distribution of errors for one variable: geopotential height at 500 hPa (zg500), 10-meter wind speed, near-surface temperature, total precipitation, Surface Downwelling Shortwave Radiation and Surface Downwelling Longwave Radiation. The error are computed for the complete test set, color bars are individually scaled per variable}}
% \label{fig:mean_error_maps}
% \end{figure}

\begin{figure}[h!]
\centering

\begin{subfigure}{\linewidth}
    \centering
    \includegraphics[width=0.98\linewidth]{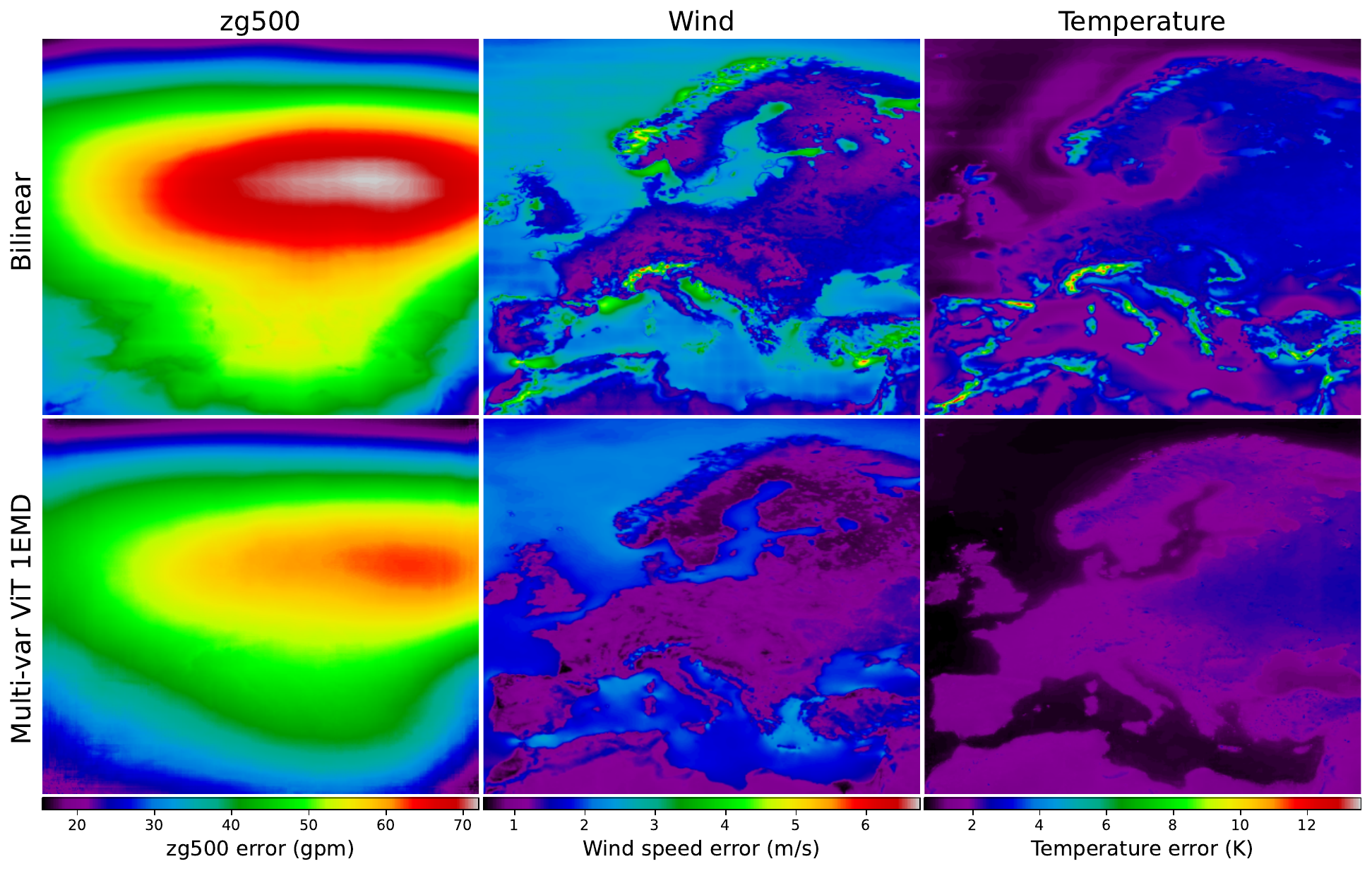}
    %\caption{}
\end{subfigure}
\begin{subfigure}{\linewidth}
    \centering
    \includegraphics[width=0.98\linewidth]{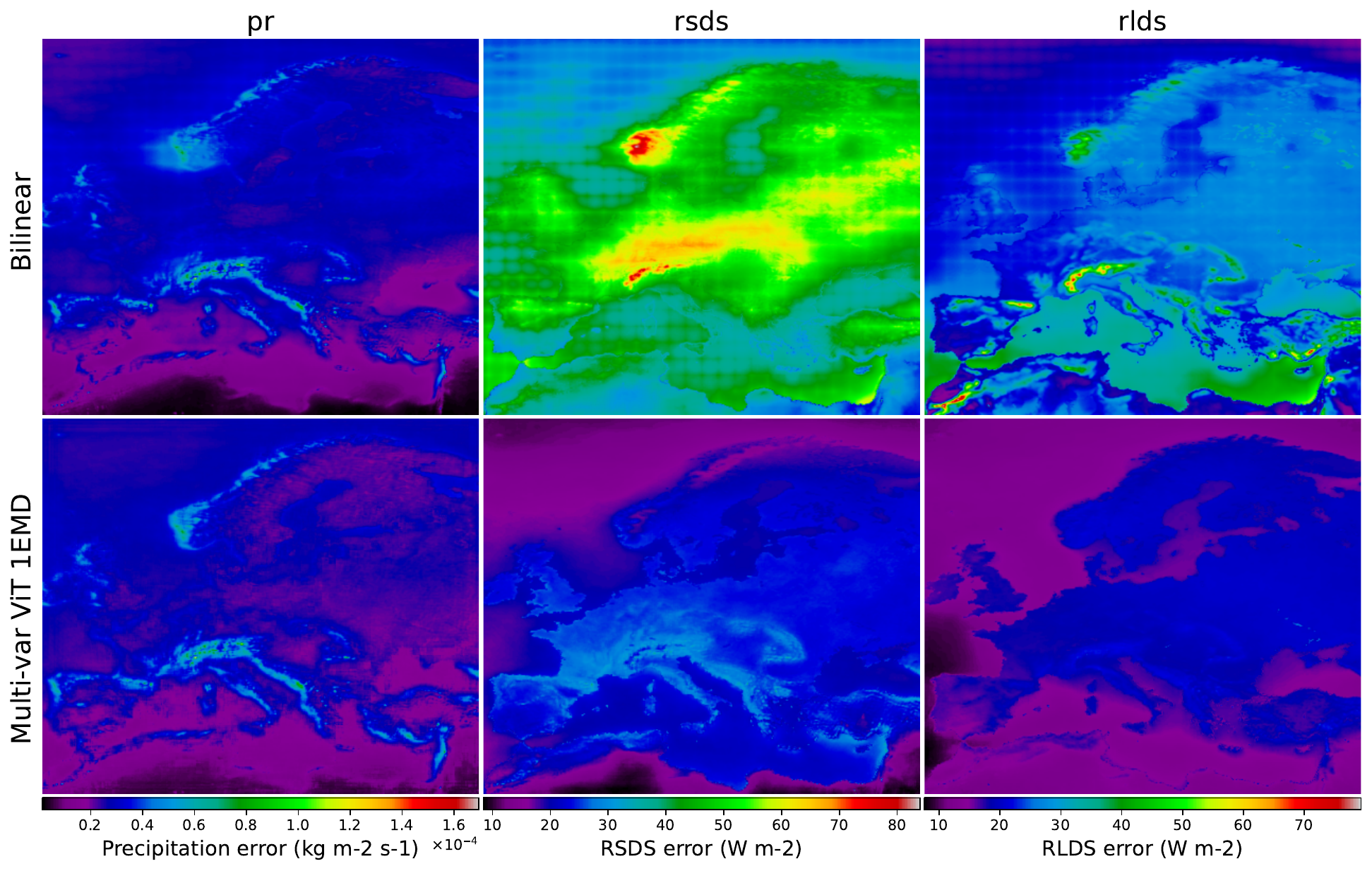}
    %\caption{}
\end{subfigure}

\caption{Mean absolute error maps for bilinear interpolation (top rows) and the 1EMD multi-variable ViT (bottom rows). The top panel shows zg500, wind speed, and temperature; the bottom panel shows precipitation, shortwave downwelling radiation, and longwave downwelling radiation. Errors are computed over the full test set, with per-variable color scales.}
\label{fig:mean_error_maps}
\end{figure}

Table~\ref{tab:variable_results_exec} reports the mean performance across two independent runs for all six variables. All deep-learning models provide substantial improvements over the bilinear interpolation baseline. For example, the MSE for near-surface temperature (tas) decreases from $9.97\times10^{-4}$ under bilinear interpolation to $4.34\times10^{-4}$ with the Single-variable ViT. Similarly, for surface wind speed (sfcWind), the bilinear error of $6.08\times10^{-3}$ is reduced by approximately half across all learned models. Comparable improvements are observed for the remaining variables, confirming the value of learning-based downscaling across a diverse set of atmospheric fields.

A second consistent pattern is that the Multi-var 1E1D ViT and the Multi-var U-Net achieve nearly identical results for all variables. For instance, on rsds the two models obtain MSE values of $5.03\times10^{-3}$ and $5.00\times10^{-3}$, while on rlds they reach $2.80\times10^{-3}$ and $2.81\times10^{-3}$, respectively. This close alignment suggests that architectural differences between transformer-based and convolutional backbones are less influential than the broader design choice of using a single shared decoder when performing multi-variable downscaling.

The comparison with the Single-variable ViT indicates that joint modeling through a single decoder results in positive transfer for several variables. Both 1E1D and the U-Net improve upon the Single-variable ViT for sfcWind (reducing MSE from $3.43\times10^{-3}$ to $3.33\times10^{-3}$ and $3.35\times10^{-3}$), for zg500 (from $1.90\times10^{-3}$ to $1.85\times10^{-3}$ and $1.84\times10^{-3}$), for rsds (from $5.37\times10^{-3}$ to $5.03\times10^{-3}$ and $5.00\times10^{-3}$), and for rlds (from $3.06\times10^{-3}$ to $2.80\times10^{-3}$ and $2.81\times10^{-3}$). These results indicate that several variables benefit from shared representation learning across related atmospheric patterns. In contrast, the Single-variable ViT performs better for tas and total precipitation (tp). For tas, the MSE increases from $4.34\times10^{-4}$ when using single-variable modeling to $4.50\times10^{-4}$ (1E1D) and $4.47\times10^{-4}$ (U-Net). For tp, the error increases from $4.46\times10^{-7}$ to $4.91\times10^{-7}$ and $5.00\times10^{-7}$. These results illustrate negative transfer for variables whose statistical or spatial characteristics are not well accommodated by a shared decoder. Additionally, the 1E1D model exhibits cross-variable leakage in zg500, with geographic artifacts emerging in the output, as illustrated in Figure \ref{fig:cross_variable_artefacts}.

The proposed 1EMD architecture achieves the best performance for all six variables. Relative to the Single-variable ViT, the MSE decreases from $4.34\times10^{-4}$ to $4.23\times10^{-4}$ for tas, from $3.43\times10^{-3}$ to $3.23\times10^{-3}$ for sfcWind, and from $4.46\times10^{-7}$ to $4.32\times10^{-7}$ for tp. The improvements are consistent across the remaining variables as well. These results indicate that combining a shared encoder with variable-specific decoders successfully incorporates the benefits of multi-variable modeling while mitigating the negative transfer observed in single-decoder approaches. To further assess the performance of 1EMD, a visual comparison of mean absolute error maps between 1EMD and bilinear interpolation is reported in Figure \ref{fig:mean_error_maps}.

Additionally, Table \ref{tab:spatiotemporal_corr} reports spatial and temporal Pearson correlations for all variables and models. Spatial correlation quantifies, at each time step, the agreement between predicted and observed spatial patterns and is summarized as mean ± standard deviation across time. Temporal correlation analogously measures, at each grid point, the agreement between predicted and observed time series and is summarized across space. Across variables, the spatial and temporal correlations reveal two broad regimes of model behaviour. For tas and zg500, all learned models perform similarly, with minimal differences between the single-variable and multivariate configurations. For rsds and rlds, learned models also converge to closely aligned performance, though a modest improvement is still observed when moving from single-variable to multivariate setups. In contrast, wind speed and total precipitation (tp) exhibit much larger performance spreads. For wind speed, the advantage of 1EMD is substantial: it clearly outperforms both the single-variable model and the other multivariate variants, achieving the highest spatial and temporal correlations (up to 0.76 and 0.64, respectively), whereas the alternative multivariate designs repeatedly underperform. Precipitation displays even more pronounced differences: most multivariate models deteriorate markedly (e.g., temporal correlation dropping to 0.32 in 1E1D), while 1EMD reaches 0.41, surpassing both the single-variable model (0.39) and all other multivariate approaches. Overall, 1EMD emerges as the most robust multivariate design, with the largest gains observed for the more challenging variables.

Multi-variable models offer clear efficiency advantages by consolidating multiple downscaling tasks into a single model, utilizing a single shared encoder thus reducing the number of computations and requiring fewer model initializations overall. To quantify this, we measure the inference time for each model to generate predictions over the 10-year test dataset. All models are run on the same hardware and share a unified pre-processing pipeline.

In Table \ref{table:efficiency} we report the number of parameters, execution time, and training time for all models. Although the 1EMD model has a substantially larger number of trainable parameters due to its six independent decoders, its inference time remains similar to that of the 1E1D model. This is expected: using six decoders instead of one increases the parameter count but does not significantly change the number of arithmetic operations, as each decoder performs the same sequence of computations with different weights. Moreover, the transformer encoder dominates the total computational cost, meaning that differences in decoder size contribute only marginally to overall runtime.

\begin{figure*}[h!]
\centering
\includegraphics[width=0.9\linewidth]{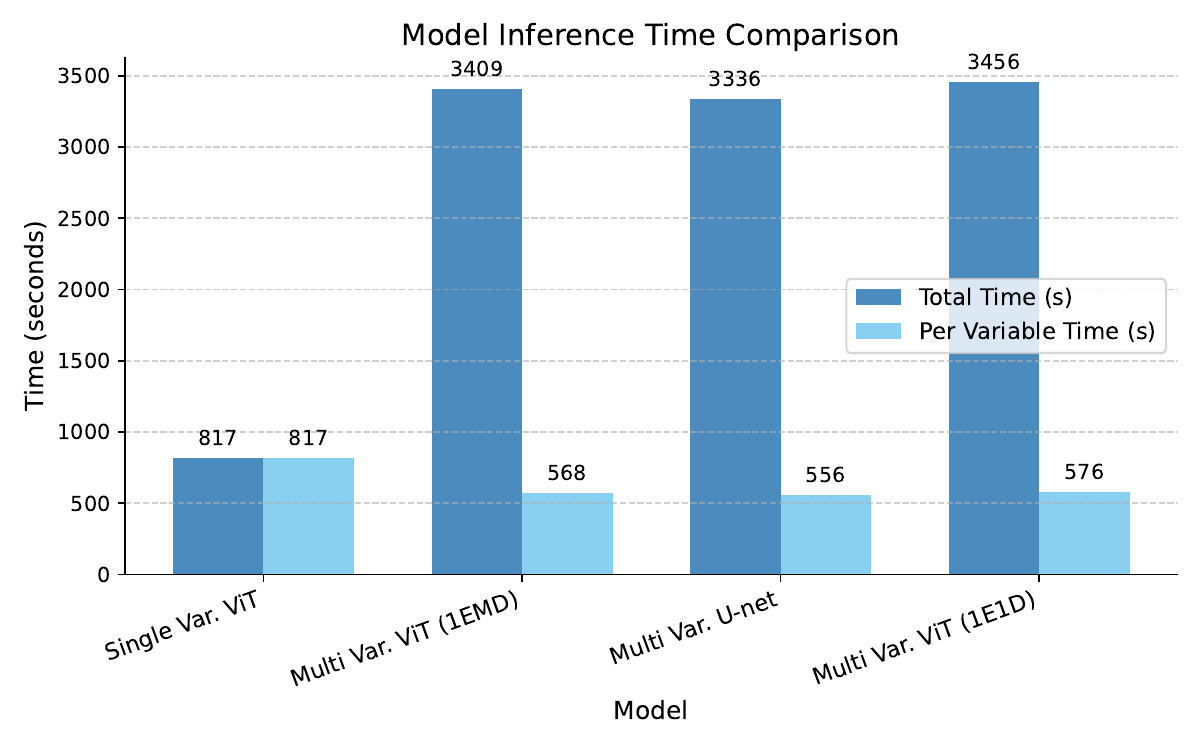}
\caption{Inference time comparison across model architectures.
The Total inference time (dark blue bars) and per-variable inference time (light blue bars) are shown for four model configurations. All models were evaluated on identical hardware (NVIDIA A100), with the same input data pipeline. The measurements correspond to the inference time required to process 10 years of daily testing data. For the multi-variable models, which predict six variables simultaneously, the per-variable inference time was calculated by dividing the total inference time the number of variables.}
\label{fig:performance-comparison}
\end{figure*}

Figure \ref{fig:performance-comparison} presents the total runtime per model (in seconds), along with per-variable time for multi-variable models which process all six variables simultaneously.

From an inference cost standpoint, executing six independent single-variable ViT models over the 10-year test period would require a cumulative runtime of $6 \times 817\,\text{s} = 4{,}902\,\text{s}$. In contrast, the multi-variable architectures complete the same six-variable downscaling task in $3{,}336$--$3{,}456\,\text{s}$. 
This corresponds to a reduction in execution time of approximately $29\%$--$32\%$, demonstrating that multi-variable processing yields a substantial improvement in inference efficiency. It is important to note that these timings exclude the additional overhead associated with loading and managing six separate models (such 
as repeated weight initialization, memory allocation, and session handling) all of which would further amplify the real-world efficiency gains of the multi-variable approach.

Training costs follow a similar pattern. Training six single-variable ViT models independently would require $6 \times 11{,}200\,\text{s} = 67{,}200\,\text{s}$, whereas the multi-variable models achieve full six-variable training in $58{,}400$--$61{,}320\,\text{s}$. This represents an overall reduction of approximately $8\%$--$13\%$ in wall-clock training time. As with inference, these estimates do not account for overheads inherent to conducting six separate training runs (such as checkpoint management, logging, scheduling, and data loading) which suggests that the effective operational efficiency of the multi-variable configurations is likely even greater in practice.

\begin{table}[htbp]
\centering
\def\arraystretch{1.1}
\begin{tabular}{lccc}
\toprule
\textbf{Model} & \textbf{Exec. Time (s)} & \textbf{Train Time (s)} & \textbf{Params (M)} \\
\midrule
Single-variable ViT     & 817   & 11\,200 & 11.63 \\
1E1D ViT                & 3\,409 & 58\,400 & 15.39 \\
Multi-variable U-Net    & 3\,336 & 61\,320 & 5.20 \\
1EMD ViT                & 3\,456 & 60\,200 & 32.25 \\
\bottomrule
\end{tabular}
\caption{Computational cost and model complexity for the three examined ViT architectures. 
Execution time is measured over the entire 10-year test dataset. Training time corresponds to 
400 training epochs under identical hardware and hyperparameter settings.}
\label{table:efficiency}
\end{table}

\section{Conclusion}

In this study, we have explored multi-variable deep learning approaches for climate downscaling, specifically targeting the GCM-to-RCM emulation task.
While most deep learning approaches focus on modeling single variables, we have argued that jointly processing multiple variables can improve both efficiency and predictive performance. Redundant modeling across variables can be avoided, and information from one variable can enhance the prediction of another by capturing inter-variable dependencies.
To test this hypothesis, we evaluated multiple deep learning architectures within a unified experimental framework, including both Vision Transformer (ViT)–based models and a multi-variable U-Net baseline. Across single-variable and multi-variable formulations, these configurations span approaches from fully independent models to joint encoding with shared or variable-specific decoders. This progression enables a systematic assessment of the trade-offs between shared representations and variable specialization, culminating in the proposed Multi Var. ViT (1EMD) model, which combines a shared encoder with dedicated decoders for each variable.
To rigorously evaluate our methods, we established a consistent and fair comparison framework. All ViT models share the same architecture, differing only in the components necessary to support multi-variable processing. Training hyperparameters, learning rate scheduling, pre-processing pipelines, and hardware resources were kept identical across all experiments. 
Our results, evaluated on a 10-year held-out test set, show that multi-variable modeling provides both accuracy and efficiency benefits over single-variable approaches. In particular, the proposed Multi Var. ViT (1EMD) consistently achieves the best performance across all variables, reducing the MSE by 5.5 \% on average compared to single-variable ViT. In addition, multi-variable models substantially reduce computational cost, resulting in 29–32\% lower inference time per variable compared to single-variable models
Overall, we have introduced an effective and straightforward method for climate downscaling in a multi-variable context. By modeling variables jointly, our results reveal a substantial gain in both performance and computational efficiency. Several open questions remain, including how many variables can be effectively modeled together, which variables should be selected to maximize performance, and whether more sophisticated loss functions could further enhance the model's capabilities.

\section*{Declarations}

\subsection*{Funding}
This research was partially funded and supported by the following Projects:
\begin{itemize}
\item European Cordis Project ``Optimal High Resolution Earth System Models for Exploring Future Climate Changes'' (OptimESM),
Grant agreement ID: 101081193
\item ISCRA Project ``AI for weather analysis and forecast'' (AIWAF)
\end{itemize}

\subsection*{Acknowledgements}
We are grateful to Thomas Noël (TCDF) for providing access to the GCM–RCM simulation data used in this study. We also thank Leone Cavicchia and Enrico Scocimarro (CMCC) for their valuable insights and constructive discussions, which helped improve the quality of this work.

% \subsection*{Authors' contributions}
% FM designed the models, implemented the experiments and analyzed the results. HL provided the climate datasets and domain expertise.

\subsection*{Statements}
The authors declare no competing interests.

\subsection*{Code and data availability}
The code is available from the authors upon reasonable request.

The data used in this study were obtained from the CORDEX archive hosted on the Earth System Grid Federation (ESGF) via the IPSL node \hyperlink{https://esgf-node.ipsl.upmc.fr/search/cordex‑ipsl/}{https://esgf-node.ipsl.upmc.fr/search/cordex‑ipsl/}

\markboth{}{}

\bibliography{bibliography}
\end{document}